\documentclass[letterpaper]{article} % DO NOT CHANGE THIS
\usepackage{aaai24}  % DO NOT CHANGE THIS
\usepackage{times}  % DO NOT CHANGE THIS
\usepackage{helvet}  % DO NOT CHANGE THIS
\usepackage{courier}  % DO NOT CHANGE THIS
\usepackage[hyphens]{url}  % DO NOT CHANGE THIS
\usepackage{graphicx} % DO NOT CHANGE THIS
\usepackage{natbib}  % DO NOT CHANGE THIS AND DO NOT ADD ANY OPTIONS TO IT
\usepackage{caption} % DO NOT CHANGE THIS AND DO NOT ADD ANY OPTIONS TO IT
\usepackage{algorithm}
\usepackage{algorithmic}

\usepackage{newfloat}
\usepackage{listings}
\DeclareCaptionStyle{ruled}{labelfont=normalfont,labelsep=colon,strut=off} % DO NOT CHANGE THIS
\floatstyle{ruled}
\newfloat{listing}{tb}{lst}{}
\floatname{listing}{Listing}
\usepackage{amsmath,amssymb}
\usepackage{array}
\usepackage{booktabs}
\usepackage{multirow}

\title{Viewpoint Integration and Registration with Vision Language Foundation Model for Image Change Understanding}
\author{
    Xiaonan Lu\textsuperscript{\rm 1,\rm 2},
    Jianlong Yuan\textsuperscript{\rm 1}\thanks{Corresponding author.},
    Ruigang Niu\textsuperscript{\rm 1},
    Yuan Hu\textsuperscript{\rm 1},
    Fan Wang\textsuperscript{\rm 1},
}
\affiliations{
    \textsuperscript{\rm 1}DAMO Academy, Alibaba Group, Beijing, China\\
    \textsuperscript{\rm 2}University of Chinese Academy of Sciences, Beijing, China\\
    \{luxiaonan.lxn, gongyuan.yjl, niuruigang.nrg, lavender.hy, fan.w\}@alibaba-inc.com
}

\usepackage{bibentry}
\begin{document}

\maketitle

\begin{abstract}
Recently, the development of pre-trained vision language foundation models (VLFMs) has led to remarkable performance in many tasks. However, these models tend to have strong single-image understanding capability but lack the ability to understand multiple images. Therefore, they cannot be directly applied to cope with image change understanding (ICU), which requires models to capture actual changes between multiple images and describe them in language. In this paper, we discover that existing VLFMs perform poorly when applied directly to ICU because of the following problems: (1) VLFMs generally learn the global representation of a single image, while ICU requires capturing nuances between multiple images. (2) The ICU performance of VLFMs is significantly affected by viewpoint variations, which is caused by the altered relationships between objects when viewpoint changes. To address these problems, we propose a Viewpoint Integration and Registration method. Concretely, we introduce a fused adapter image encoder that fine-tunes pre-trained encoders by inserting designed trainable adapters and fused adapters, to effectively capture nuances between image pairs. Additionally, a viewpoint registration flow and a semantic emphasizing module are designed to reduce the performance degradation caused by viewpoint variations in the visual and semantic space, respectively. Experimental results on CLEVR-Change and Spot-the-Diff demonstrate that our method achieves state-of-the-art performance in all metrics.
\end{abstract}

\begin{figure}[!t]
\centering
\includegraphics[width=0.95\columnwidth]{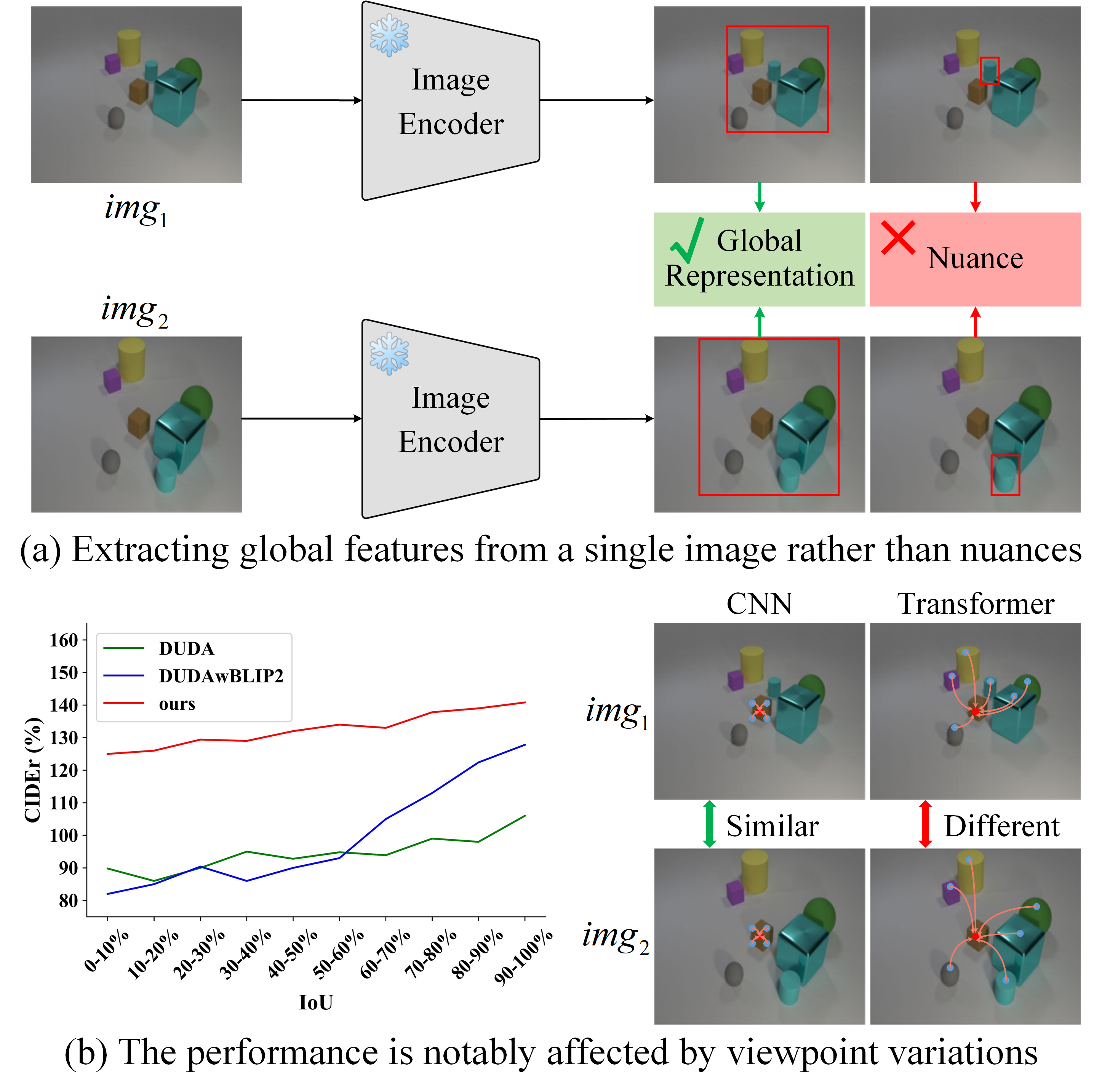} % Reduce the figure size so that it is slightly narrower than the column.
\caption{Challenges of directly applying VLFMs to ICU. (a) Pre-trained VLFMs only extract global representation of a single image rather than nuance features between two images. (b) When there are viewpoint variations, the relationships between objects are altered, and the global self-attention in VLFMs leads to obviously different features of the same objects, resulting in performance degradation.}
% Our method improves the performance through viewpoint integration and registration.
\label{fig1}
\end{figure}

\section{Introduction}

With the advent of transformer \cite{vaswani2017attention} and large-scale datasets \cite{schuhmann2021laion,kirillov2023segment}, numerous foundation models of vision \cite{xie2022simmim,fang2023eva} and language \cite{touvron2023llama,chiang2023vicuna} have emerged. They are pre-trained on abundant data and acquire generalized knowledge. Based on the success of single-modal foundation models, there has been growing interest in multi-modal vision language foundation models (VLFMs) \cite{li2023blip,dai2023instructblip} which have demonstrated advanced performance in tasks such as single image caption and visual question answering.

Nevertheless, VLFMs, which typically comprise an image encoder and a large-scale language model (LLM), tend to extract features from each image independently without considering the viewpoint differences between multiple images. Thus, these models excel at global feature extraction and understanding for individual images but lack the ability to extract relationships between multiple images and accurately comprehend them, particularly in cases where viewpoint changes. It is important for image change understanding (ICU) \cite{jhamtani2018learning,park2019robust}, which requires models to describe the actual nuances between images while excluding extraneous viewpoint variations. Compared to image change detection \cite{radke2005image,chen2023saras} and image caption \cite{xu2015show,li2017image}, ICU is more complex and has broad applications in fields such as facility monitoring, medical imaging, and aerial photography \cite{tu2021semantic}.

Existing VLFMs perform poorly in ICU due to their inability to capture nuances between images and their sensitivity to viewpoint variations. Firstly, most ICU methods and VLFMs typically load pre-trained image encoders and freeze their parameters, only training subsequent modules for image-text alignment and sentence generation. As Figure~\ref{fig1}(a) shows, VLFMs only learn the global representation of a single image and cannot capture nuances between two images. And fine-tuning the entire network would require plentiful multi-image data to avoid over-fitting and consume significant computing resources and time. Secondly, based on a conventional ICU method DUDA \cite{park2019robust}, we replace its image encoder and language decoder with BLIP-2 \cite{li2023blip} pre-trained image encoder and LLM and freeze their parameters while keeping the other modules. As shown in Figure~\ref{fig1}(b), the performance of the modified model declines more sharply as the change of viewpoint increases, \textit{i.e.} IoU decreases. This is because the self-attention in VLFMs aggregates features similar to the current position across the entire image, while CNN only aggregates contextual features around the current position. When there are viewpoint variations, the relationships between objects change, and the features extracted by VLFMs have more noticeable differences of the same objects. Hence, VLFMs are sensitive to viewpoint variations and cannot determine whether the differences are caused by actual changes or viewpoint variations, resulting in incorrect identification of changed regions and inaccurate captions.

To address the aforementioned issues, we propose a novel Viewpoint Integration and Registration method with VLFMs for ICU, dubbed VIR-VLFM. It models the correlation between images through preliminary integration in the designed image encoder, and then performs viewpoint registration and semantic enhancement in the visual and language space, respectively, to highlight nuance features for LLM to generate change captions. Specifically, based on the idea of adapter tuning \cite{houlsby2019parameter,yang2023aim}, a fused adapter image encoder is proposed to equip pre-trained image encoders with the ability to capture nuances between images, which is achieved by inserting trainable adapters and fused adapters in transformer blocks of the image encoder. This module adapts the pre-trained image encoder to extract nuances between two images and bridges the gap between image encoder pre-training and ICU tasks efficiently with only minimal extra trainable parameters. Furthermore, to address the severe performance degradation resulting from viewpoint variations, we propose a viewpoint registration flow and a semantic emphasizing module. The former module, inspired by optical flow \cite{houlsby2019parameter}, aligns the features of two images in the visual space, reducing the interference of viewpoint variations. The latter module further highlights the features that have actually changed in the semantic space, and then sends them to LLM for generating captions. Overall, our method endows VLFMs with the ability to understand multiple images through the proposed modules, surpassing existing state-of-the-art ICU methods.

The major contributions of this work are as follows:
\begin{itemize}
    \item We propose a novel method VIR-VLFM, which is the first attempt to enhance the multi-image understanding ability of VLFMs, enabling them to be applied to ICU.
    \item A fused adapter image encoder is devised to bridge the gap between image encoder pre-training and ICU. Besides, a viewpoint registration flow and a semantic emphasizing module are designed to reduce the severe performance degradation caused by viewpoint variations.
    \item Extensive experiments on CLEVR-Change and Spot-the-Diff illustrate that our method achieves state-of-the-art performance in image change caption on all metrics and shows promising results in change question answering.
\end{itemize}

\begin{figure*}[t]
\centering
\includegraphics[width=\textwidth]{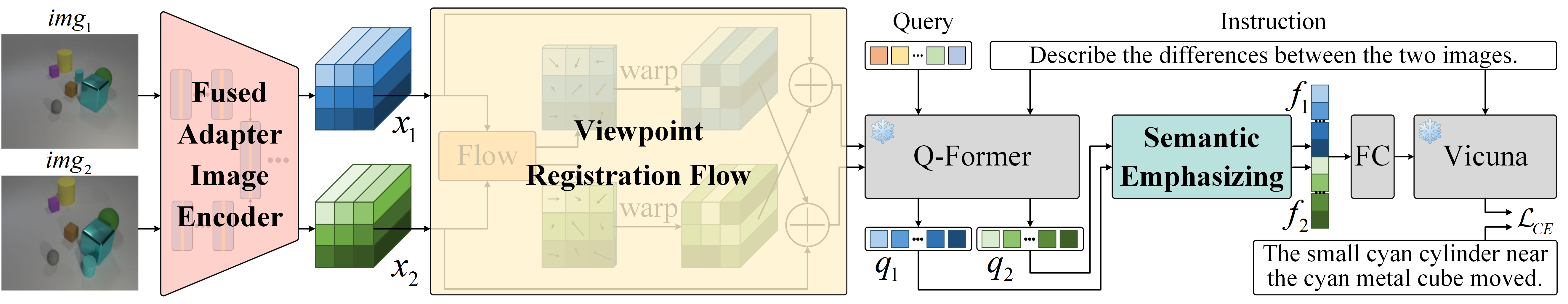} % Reduce the figure size so that it is slightly narrower than the column.
\caption{Architecture of our proposed VIR-VLFM. Based on the BLIP-2 pre-trained image encoder, including a ViT backbone and a Q-Former, and the Vicuna LLM, our method proposes a fused adapter image encoder that introduces adapters and fused adapters to integrate the two input images. Besides, a viewpoint registration flow and a semantic emphasizing module are devised to achieve viewpoint alignment and nuance emphasis in the visual space and semantic space, respectively.}
\label{fig2}
\end{figure*}

\section{Related Work}

\subsubsection{Image Change Understanding.}
ICU combines image change detection and image caption to generate descriptions based on the differences between two images. DDLA \cite{jhamtani2018learning} first collects a dataset from video-surveillance footage and proposes a model that captures visual salience by aligning pixel clusters with sentences. To address viewpoint changes, DUDA \cite{park2019robust} constructs a synthetic dataset called CLEVR-Change and proposes a dual-attention to locate changed region and a dynamic speaker to select features for prediction. It directly applies the subtraction of features, which cannot accurately capture actual changes with viewpoint variations. M-VAM \cite{shi2020finding} proposes a viewpoint-agnostic encoder to exhaustively measure feature similarities across different regions. And $\rm R^3Net$ \cite{tu2021r} devises a relation-embedded module and a representation reconstruction module to explicitly model change representation. IDC \cite{yao2022image} designs three self-supervised tasks to align visual differences with descriptions at a fine-grained level. CLIP4IDC \cite{guo2022clip4idc} introduces CLIP \cite{radford2021learning} and an adaptation training process to capture nuances. NCT \cite{tu2023neighborhood} proposes a neighboring feature aggregating module to locate inconspicuous changes. Additionally, there are also some methods proposed for remote sensing \cite{liu2023progressive,chang2023changes}.

\subsubsection{Vision Language Foundation Model.}
Benefiting from the rapid development of LLMs, several vision language foundation models are proposed to integrate visual information with LLMs. Flamingo \cite{alayrac2022flamingo} introduces a gated cross-attention to align the features of two modalities. BLIP-2 \cite{li2023blip} designs a Querying Transformer (Q-Former) that employs learnable queries to transform visual features from an frozen image encoder. It acts as an information bottleneck between the frozen image encoder and the LLM to achieve effective vision-language alignment. InstructBLIP \cite{dai2023instructblip} collects 26 publicly datasets and proposes an instruction fine-tuning paradigm for training BLIP-2 to improve zero-shot performance. Additionally, LLaVA \cite{liu2023visual} and MiniGPT-4 \cite{zhu2023minigpt} achieve notable advancements by fine-tuning the fully connected layer and Q-Former on meticulously collected data. mPLUG-Owl \cite{ye2023mplug} employs low-rank adaptation \cite{hu2021lora} to jointly train the visual encoder and LLM.

However, existing VLFMs mainly focus on captioning and multi-turn question answering for single image and cannot capture relationships between multiple images. Our method is proposed to enables VLFMs to understand multiple images and be applied for ICU tasks.

\section{Method}

% In this section, we first briefly introduce the architecture and the training strategy of our proposed VIR-VLFM. Then, we separately elaborate on the devised modules.

\begin{figure*}[t]
\centering
\includegraphics[width=0.96\textwidth]{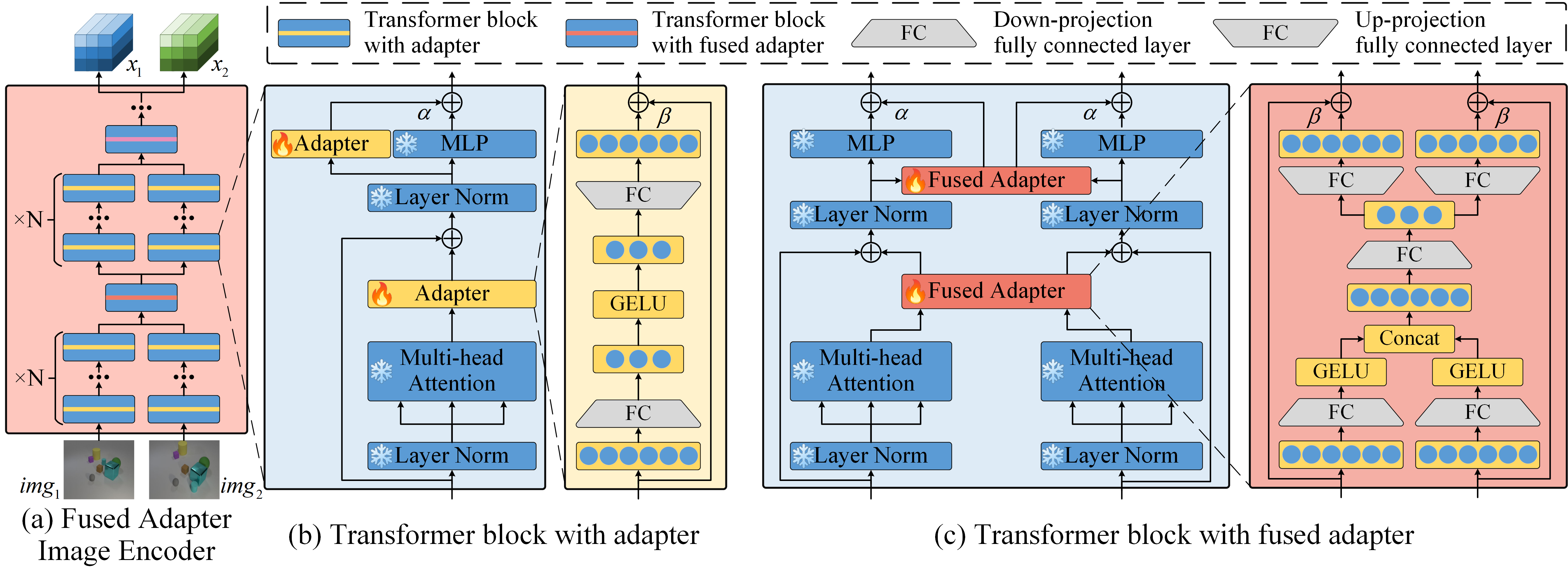} % Reduce the figure size so that it is slightly narrower than the column.
\caption{Architecture of the fused adapter image encoder (a). It consists of transformer blocks with embedded adapters (b) and fused adapters (c). The block with fused adapter is applied every $N$ blocks to integrate two features.}
\label{fig3}
\end{figure*}

\subsection{Overview}

We devote to embed VLFMs for ICU and propose a Viewpoint Integration and Registration method based on BLIP-2 \cite{li2023blip}, dubbed VIR-VLFM. Relying on the pre-trained visual encoder and LLM to provide high-quality visual representations and powerful language generation, BLIP-2 introduces a Q-Former between the two models to bridge the modality gap. And through a two-stage training strategy, it achieves state-of-the-art performance on various visual language tasks. However, like most other VLFMs, BLIP-2 still has limited ability to understand multiple images. Thus, we choose BLIP-2 as the foundation for our model construction and enhance its ability to understand multiple images through viewpoint integration and registration, making it applicable for ICU tasks.

As Figure~\ref{fig2} shows, our VIR-VLFM leverages EVA-ViT-g/14 \cite{fang2023eva} applied in BLIP-2 as the original image encoder and Vicuna \cite{chiang2023vicuna} as the language decoder. A Q-Former and a fully connected layer pre-trained by BLIP-2 are also introduced to eliminate the modality gap. Building upon this, VIR-VLFM introduces a fused adapter image encoder, a viewpoint registration flow, and a semantic emphasizing module to construct a model with the ability to understand multiple images. Given two input images, $img_{i} \in \mathbb{R}^{ H \times W \times 3},~i\in \left \{ 1,2 \right \}$, the image encoder first embeds them into $n$ image patches as follows:
\begin{equation}
	\label{Eq 1}
	x_{i}=\left \{ img_{i}^{cls},img_{i}^{1},\cdots ,img_{i}^{n}  \right \}+p,~i\in \left \{ 1,2 \right \}
\end{equation}
where $img_{i}^{j} \in \mathbb{R}^{d}$ and $d$ is the dimension of patch. $n=\frac{H}{r} \times \frac{W}{r}$ and $r$ is the size of patch. $img_{i}^{cls} \in \mathbb{R}^{d}$ is a learned class embedding and $p \in \mathbb{R}^{d}$ represents the positional embedding.

The fused adapter image encoder introduces adapters and fused adapters in transformer blocks to preliminarily integrate the two images, obtaining visual features $x_{1}$ and $x_{2}$. Then, they are aligned through the viewpoint registration flow, and the registered visual features are transformed by the Q-Former to generate semantic features $q_{1}$ and $q_{2}$. The semantic emphasizing module further highlights $q_{1}$ and $q_{2}$ to emphasize nuance features in the semantic space, obtaining $f_{1}$ and $f_{2}$. The two features are then concatenated and fed into the fully connected layer to map the dimensions same as the input of LLM. Finally, the features are combined with the instruction "Describe the differences between the two images." and sent to LLM for predicting change captions.

During training, the original layers in the image encoder, Q-Former, and LLM are frozen. Only the adapters in the fused adapter image encoder, viewpoint registration flow, semantic emphasizing module, and fully connected layer are trained. The training loss is a word-wise cross-entropy loss.
\begin{equation}
	\label{Eq 2}
	\mathcal L_{CE}=-\sum_{t=1}^{m}w_{t}^{g}\log p_{\theta }\left ( w_{t}^{p} | w_{1:t-1}^{p}, img_{1}, img_{2} \right ) 
\end{equation}
where $m$ is the length of caption, and $\theta$ is the trainable parameters. $w_{t}^{g}$ and $w_{t}^{p}$ are one-hot encoding of the $t$-th word in the ground-truth and the predicted caption, respectively.

\subsection{Fused Adapter Image Encoder}

The fused adapter image encoder introduces trainable adapters and fused adapters to adapt pre-trained image encoders to ICU. As Figure~\ref{fig3}(a) shows, we insert adapters in each transformer block to extract appropriate features of each image. And a fused adapter is applied to replace adapters every $N$ blocks for integrating two features, promoting the encoder to extract nuances between two images.

As Figure~\ref{fig3}(b) shows, two adapters are inserted in each block, one is added after the multi-head self-attention (MSA), and the other is in parallel with the multi-layer perceptron (MLP). Similar to the adapters used for natural language processing, the introduced adapter is a bottleneck structure including a down-projection, an activation function, and an up-projection, which is formulated as follows:
\begin{equation}
    \label{Eq 3}
    x_{i}^{adp}=F^{U}\left ( GELU\left ( F^{D}\left ( x_{i} \right )   \right )  \right ) + \beta x_{i}
\end{equation}
where $F^{U}$ and $F^{D}$ are the up- and down-projections, and $GELU$ is a non-linear activation function. Thus, the transformer block with adapter can be formulated as follows:
\begin{equation}
    \label{Eq 4}
    \tilde{x}_{i} =LN\left ( Adp_{1} \left ( MSA\left ( LN\left ( x_{i} \right )  \right )  \right ) + x_{i}  \right )
\end{equation}
\begin{equation}
    \label{Eq 5}
    x_{i}=MLP\left ( \tilde{x}_{i}  \right ) + \alpha Adp_{2}\left ( \tilde{x}_{i} \right ) 
\end{equation}
where $LN$ is layer norm. $Adp_{1}$ and $Adp_{2}$ represent adapters after MSA and parallel with MLP, respectively. $\alpha$ is set to $0.5$, and $\beta$ in $Adp_{1}$ and $Adp_{2}$ is set to $1$ and $0$.

Additionally, ICU requires the image encoder integrating two images to capture nuances. Hence, a fused adapter is designed to replace adapters in several transformer blocks. As shown in Figure~\ref{fig3}(c), the fused adapter is a dual-input and dual-output module. First, it reduces the dimension of two features through two down-projection layers followed by GELU activation. Then, it concatenates two features and integrates them via a projection layer. After that, the integrated feature is fed into two up-projection layers to obtain features with the same dimension as the input. The features are added to their respective original features to obtain output features. The fused adapter is formulated as follows:
\begin{equation}
    \label{Eq 6}
    x_{1,2}=F\left ( \left [ GELU\left ( F_{1}^{D}\left ( x_{1} \right ) \right )|| GELU\left ( F_{2}^{D}\left ( x_{2} \right ) \right ) \right ]  \right )
\end{equation}
\begin{equation}
    \label{Eq 7}
    x_{i}^{fused\_adp}=F_{i}^{U}\left (  x_{1,2} \right )+\beta x_{i}, i\in \left \{ 1,2 \right \}
\end{equation}
where $\left [ \cdot||\cdot \right ]$ means concatenation. The inserted positions of fused adapters in transformer blocks and the hyper-parameters $\alpha$ and $\beta$ are the same as adapters. Only the parameters of the newly embedded adapters and fused adapters are trained, while the remaining layers are loaded from the pre-trained weights and kept frozen. The parameters of $F^{U}$ are initialized to 0 to ensure that the output of the pre-trained encoder is not altered in the initial training phase.

\begin{figure}[t]
\centering
\includegraphics[width=0.93\columnwidth]{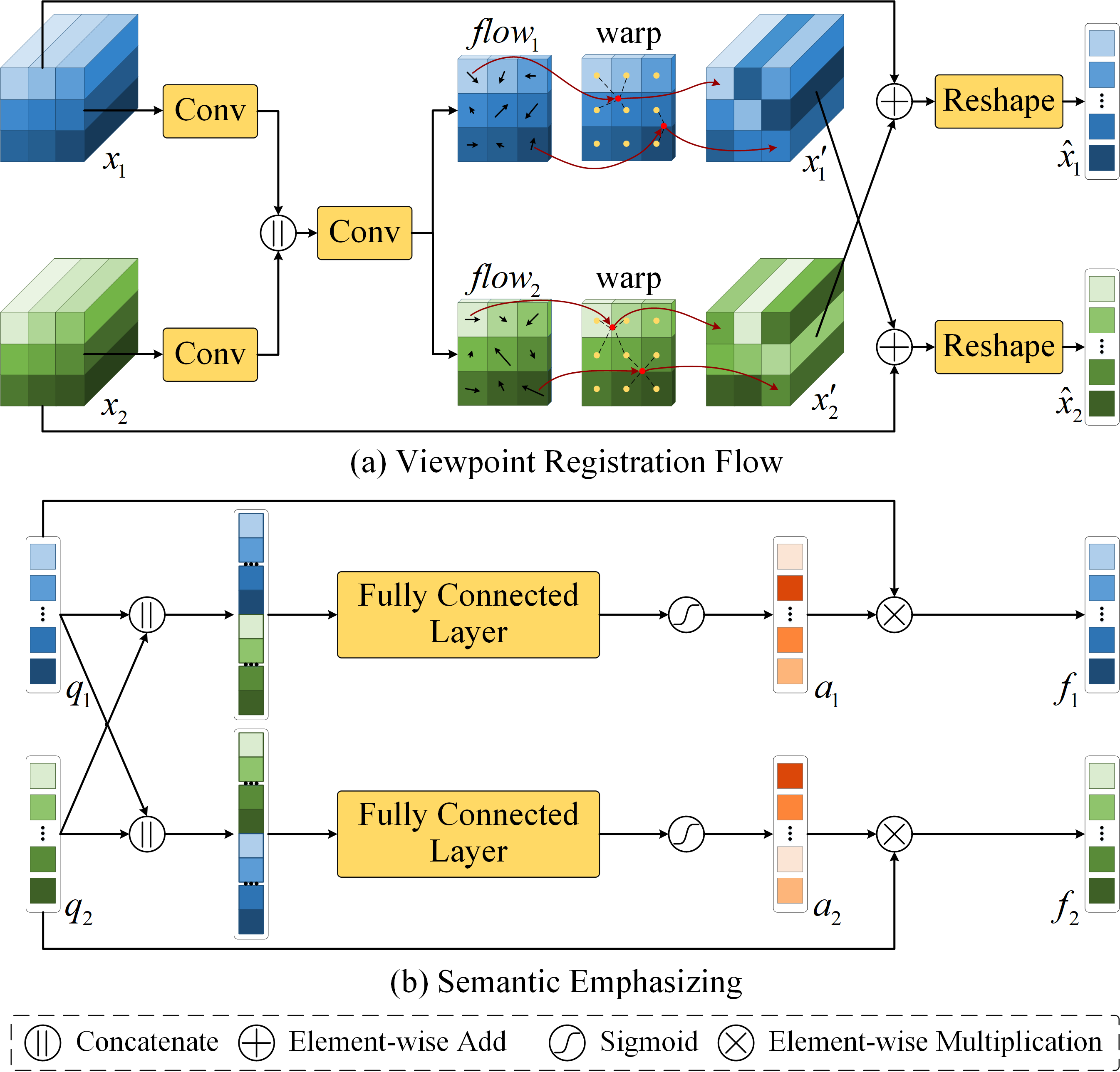} % Reduce the figure size so that it is slightly narrower than the column.
\caption{Architecture of (a) the viewpoint registration flow and (b) the semantic emphasizing module.}
\label{fig4}
\end{figure}

\begin{table*}[t]
	\setlength{\tabcolsep}{4.6pt}
	\renewcommand{\arraystretch}{1.2}
	\begin{center}
		\begin{tabular}{c|ccccc|ccccc|ccccc}
			\toprule
			\multirow{2}{*}{Method} & \multicolumn{5}{c|}{Total} & \multicolumn{5}{c|}{Scene Change} & \multicolumn{5}{c}{Distractor} \\ 
			& B & C & M & S & R & B & C & M & S & R & B & C & M & S & R \\ 
			\hline
			DUDA \shortcite{park2019robust} & 47.3 & 112.3 & 33.9 & 24.5 & - & 42.9 & 94.6 & 29.7 & 19.9 & - & 59.8 & 110.8 & 45.2 & 29.1 & - \\
                DUDA w BLIP-2 & 45.9 & 113.1 & 33.8 & 25.0 & 67.0 & 41.8 & 110.7 & 28.6 & 19.0 & 61.6 & 48.5 & 118.3 & 45.8 & 31.0 & 72.4 \\ 
                \hline
                +FAIE & 54.1 & 142.9 & 39.8 & 30.3 & 74.6 & 50.5 & 130.5 & 34.6 & 28.6 & 72.9 & 58.4 & 137.8 & 47.6 & 31.9 & 75.5 \\ 
                +VRF & 50.1 & 130.0 & 37.5 & 28.9 & 71.2 & 44.4 & 121.8 & 32.5 & 22.1 & 67.9 & 59.3 & 134.2 & 47.4 & 32.5 & 74.9 \\ 
                +SEM & 47.2 & 117.8 & 35.4 & 26.9 & 69.0 & 44.4 & 114.5 & 31.1 & 23.7 & 64.9 & 49.2 & 123.3 & 46.1 & 31.3 & 72.9 \\ 
                \hline
                +FAIE+VRF & 57.6 & 149.3 & 41.2 & 33.3 & 77.1 & 55.2 & 136.2 & 38.1 & 30.0 & 76.8 & 63.2 & 144.9 & 51.3 & 33.7 & 78.1 \\
                +FAIE+VRF+SEM & \textbf{58.2} & \textbf{153.4} & \textbf{42.6} & \textbf{34.5} & \textbf{78.9} & \textbf{57.1} & \textbf{136.7} & \textbf{39.7} & \textbf{32.5} & \textbf{77.7} & \textbf{67.9} & \textbf{149.2} & \textbf{53.5} & \textbf{36.8} & \textbf{79.8} \\
			\bottomrule
		\end{tabular}
		\caption{Ablation study of different modules. FAIE means the fused adapter image encoder. VRF means the viewpoint registration flow. SEM means the semantic emphasizing module. Best results are in \textbf{bold}.}
		\label{Table 1}
	\end{center}
\end{table*}

\subsection{Viewpoint Registration Flow}

Based on the optical flow \cite{dosovitskiy2015flownet,ilg2017flownet,zhu2017deep} designed to align features between adjacent frames in video processing tasks, Li \textit{et al.} \shortcite{li2020semantic} design a semantic flow to align features between adjacent levels of feature pyramid networks \cite{lin2017feature}. Drawing inspiration from the optical flow and the semantic flow, we design a viewpoint registration flow for flexible viewpoint registration between two images, which eliminates the impact of viewpoint changes on ICU performance.

To better utilize the spatial information in the viewpoint registration flow, for the features extracted from the fused adapter image encoder, $x_{i}=\left \{ x_{i}^{cls},x_{i}^{1},\cdots ,x_{i}^{n}  \right \}, i\in \left \{ 1,2 \right \}$, we first remove the class embedding $x_{i}^{cls}$ and convert them into a two-dimensional arrangement format as follows:
\begin{equation}
    \label{Eq 8}
    x_{i}\in \mathbb{R}^{\left ( n+1 \right ) \times d} \xrightarrow{remove~x_{i}^{cls}} \mathbb{R}^{n \times d}\xrightarrow{reshape} \mathbb{R}^{\frac{H}{r} \times \frac{W}{r}  \times d}
\end{equation}

As Figure~\ref{fig4}(a) shows, given two converted features, we first perform dimensionality reduction through two $1 \times 1$ convolutional layers and concatenate them together. The concatenated feature is then fed into a $3 \times 3$ convolutional layer to predict viewpoint registration flow fields $flow_{i}\in \mathbb{R}^{\frac{H}{r} \times \frac{W}{r}  \times 2}$. This process can be formulated as follows:
\begin{equation}
    \label{Eq 9}
    \left \{ flow_{1},flow_{2} \right \} =Conv\left ( \left [ Conv_{1}\left ( x_{1} \right )||Conv_{2}\left ( x_{2}\right )  \right ]  \right ) 
\end{equation}

After obtaining $\left \{ flow_{1},flow_{2} \right \}$ for two images, the differentiable bi-linear sampling mechanism \cite{jaderberg2015spatial} is used to estimate the final value at each position of the spatial grid by linearly interpolating the values of the 4-neighboring pixels (top-left, top-right, bottom-left, and bottom-right). The formula is as follows:
\begin{equation}
    \label{Eq 10}
    x'_{i}\left ( m,n \right )  = \sum_{\left ( m',n' \right )\in \mathcal{N}\left ( m,n \right ) }w\left ( m',n' \right )\cdot x_{i}\left ( m',n' \right ) 
\end{equation}
where $\mathcal{N}\left ( m,n \right )$ means four neighbors of the point $\left ( m,n \right )$. $w\left ( m',n' \right )$ denotes the bi-linear kernel weights estimated by the distance of warped grid. The viewpoints of wrapped features $x'_{1}$ and $x'_{2}$ are aligned with $x_{2}$ and $x_{1}$, respectively.

Most image change detection methods and ICU methods usually apply the subtraction between two features as the input for subsequent decoders. However, for ICU tasks, it is not only necessary to identify changed regions, but also to describe the relationship between changed regions and their surroundings before and after change. Taking subtraction between two viewpoint-aligned features will result in the vanishing of context information that has not changed. Therefore, we use the sum operation of aligned features and reshape them back to one-dimensional sequence format. It can be formulated as follows:
\begin{equation}
    \label{Eq 11}
    \left\{\begin{matrix}
    \hat{x}_{1}=reshape\left ( x_{1} + x'_{2} \right ) \\
    \hat{x}_{2}=reshape\left ( x_{2} + x'_{1} \right ) 
\end{matrix}\right.
\end{equation}
where $\hat{x}_{i} \in \mathbb{R}^{n \times d}$. Then, we concatenate $\hat{x}_{i}$ with the class embedding $x_{i}^{cls}$ as the input for the subsequent Q-Former.

\subsection{Semantic Emphasizing}

Viewpoint registration flow models viewpoint variations between images to register viewpoints in the visual space. Then, the pre-trained Q-Former transforms the aligned visual features into the semantic space by introducing learnable queries. As shown in Figure~\ref{fig4}(b), a semantic emphasizing module is designed to further enhance the nuance features in the semantic space. The module takes the transformed semantic features $q_{1}$ and $q_{2}$ as input, and concatenates them in a "self-then-other" manner. Then, the concated features are fed into a fully connected layer followed by $Sigmoid$ to obtain attention weights $a_{i}$, which is formulated as follows:

\begin{equation}
    \label{Eq 12}
    \left\{\begin{matrix}
    a_{1}=\sigma \left (  FC\left ( \left [ q_{1}||q_{2} \right ] \right ) \right ) \\
    a_{2}=\sigma \left ( FC\left ( \left [ q_{2}||q_{1} \right ] \right )  \right ) 
\end{matrix}\right.
\end{equation}
where $\sigma$ means $Sigmoid$ function and $FC$ represents a fully connected layer. $a_{i}$ is used as an attention mechanism to highlight the nuance features of $q_{i}$ that need to be described. The generated features $f_{i}$ adequately capture the characteristics of changed objects and their surrounding context, which are then fed into a fully connected layer and a LLM to generate final descriptions.
% Through the initial integration in the fused adapter image encoder, explicit viewpoint alignment in the visual space using viewpoint registration flow, and emphasis of nuance features in the semantic space using the semantic enhancement module,

\section{Experiments}

\subsection{Experimental Setup}

\subsubsection{Dataset.}

Our method is evaluated on two standard datasets for image change caption, including CLEVR-Change \cite{park2019robust} and Spot-the-Diff \cite{jhamtani2018learning}. We use the official splits same as DUDA \cite{park2019robust}. Detailed introductions are presented in the supplementary material.

\subsubsection{Evaluation Metrics.}

Following previous works, five standard metrics are applied to evaluate our method, including BLEU-4 (B) \cite{papineni2002bleu}, CIDEr (C) \cite{vedantam2015cider}, METEOR (M) \cite{banerjee2005meteor}, SPICE (S) \cite{anderson2016spice}, and ROUGE-L (R) \cite{lin2004rouge}. And we employ vocabulary ranking following InstructBLIP. Readers can refer to the supplementary material and \cite{dai2023instructblip} for more details.

\subsubsection{Implementation Details.}

The architecture of our model is roughly consistent with BLIP-2 \cite{li2023blip}. We apply EVA-ViT-g/14 \cite{fang2023eva} and Vicuna-7B \cite{chiang2023vicuna} as the image encoder and LLM, respectively. The two modules are augmented with a Q-Former and a fully connected layer pre-trained by BLIP-2. Please refer to the supplementary material and \cite{park2019robust} for more training details.

\begin{table}[t]
	\setlength{\tabcolsep}{5.0pt}
	\renewcommand{\arraystretch}{1.2}
	\begin{center}
		\begin{tabular}{c|c|ccccc}
			\toprule
			Method & $N$ & B & C & M & S & R \\ 
			\hline
                 DUDA w BLIP-2 & - & 45.9 & 113.1 & 33.8 & 25.0 & 67.0 \\
                 \hline
                 & - & 50.5 & 133.4 & 36.8 & 27.3 & 70.1 \\
                 & 10 & 52.8 & 137.2 & 38.6 & 28.9 & 73.1 \\ 
                 & 5 & 54.1 & \textbf{142.9} & \textbf{39.8} & \textbf{30.3} & \textbf{74.6} \\
                 +FAIE & 3 & \textbf{54.3} & 141.1 & 39.7 & 29.9 & 73.5 \\
                 & 2 & 54.1 & 138.5 & 39.0 & 30.2 & 73.3 \\
                 & 1 & 53.5 & 139.4 & 38.5 & 28.4 & 71.9 \\
                 & 0 & 53.0 & 137.3 & 38.7 & 28.4 & 71.1 \\
			\bottomrule
		\end{tabular}
		\caption{Ablation study of the number of interval blocks for fused adapters $N$. - indicates that no fused adapter is used.}
		\label{Table 2}
	\end{center}
\end{table}

\subsection{Ablation Study}

To analyze the effectiveness of the proposed modules, we conduct comprehensive ablations on the CLEVR-Change.

\subsubsection{Effects of Different Modules.}
The results presented in Table~\ref{Table 1} demonstrate the effectiveness of our proposed modules in VIR-VLFM. We first embed BLIP-2 \cite{li2023blip} into the ICU method DUDA \cite{park2019robust} and freeze the pre-trained parameters, named DUDA w BLIP-2. Notably, its performance is comparable to DUDA, which implies that directly applying VLFMs to ICU tasks is insufficient in adequately utilizing the knowledge learned from large-scale datasets and obtaining performance improvement. With the utilization of the fused adapter image encoder, the performance is significantly improved. This indicates that our designed adapters and fused adapters can efficiently adapt the image encoder to capture nuances between image pairs. Moreover, when individually adding the viewpoint registration flow and the semantic emphasizing module, the performance is also improved compared to DUDA w BLIP-2, especially in the case of distractors. These two modules aid in effectively eliminating the performance degradation caused by viewpoint changes. Furthermore, the combination of the three modules yields the best performance across all scenarios on all metrics, which significantly outperforms DUDA w BLIP-2.

\begin{figure}[t]
\centering
\includegraphics[width=\columnwidth]{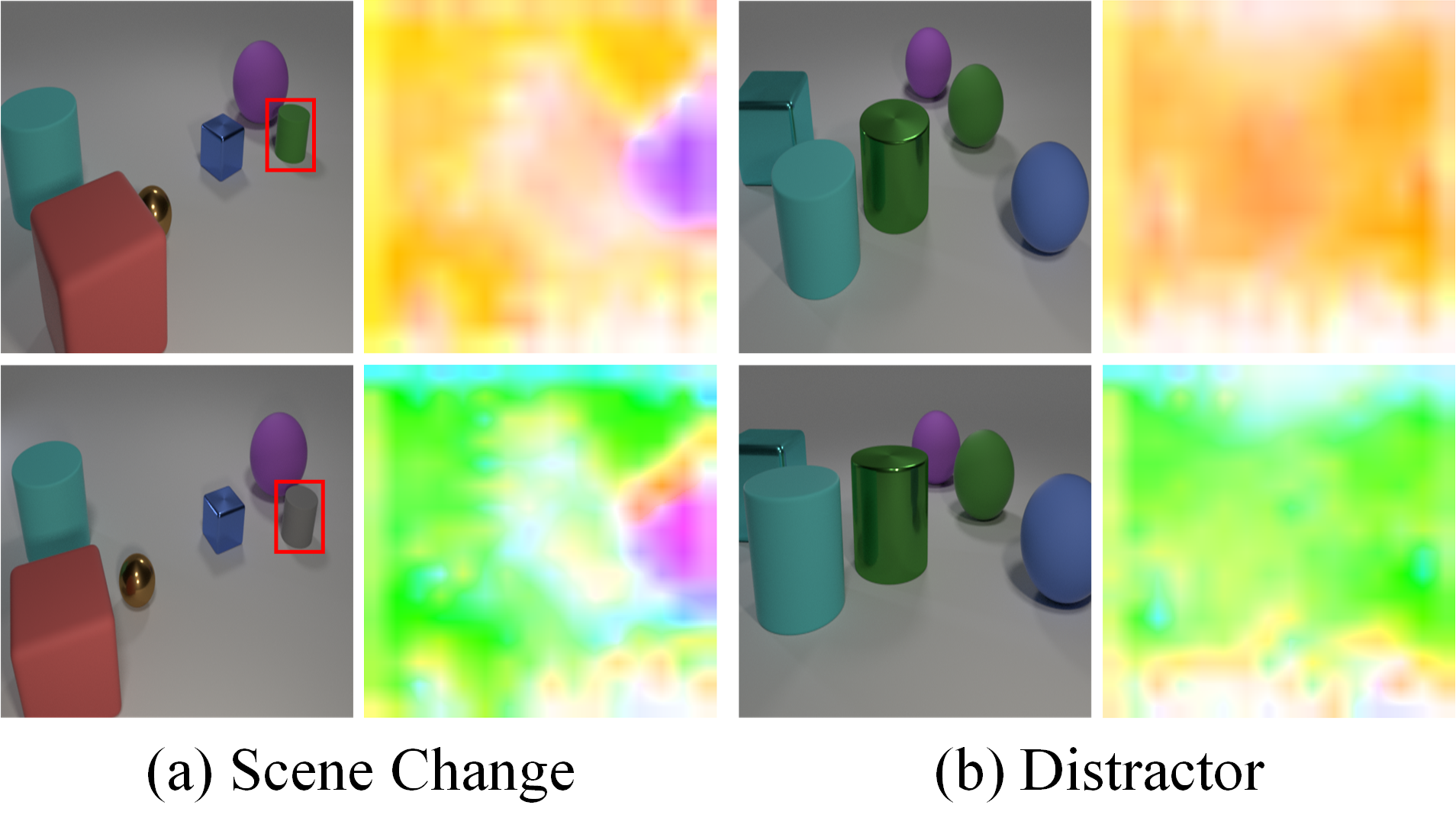} % Reduce the figure size so that it is slightly narrower than the column.
\caption{Visualization of viewpoint registration flow fields.}
\label{fig5}
\end{figure}
\begin{figure}[t]
\centering
\includegraphics[width=\columnwidth]{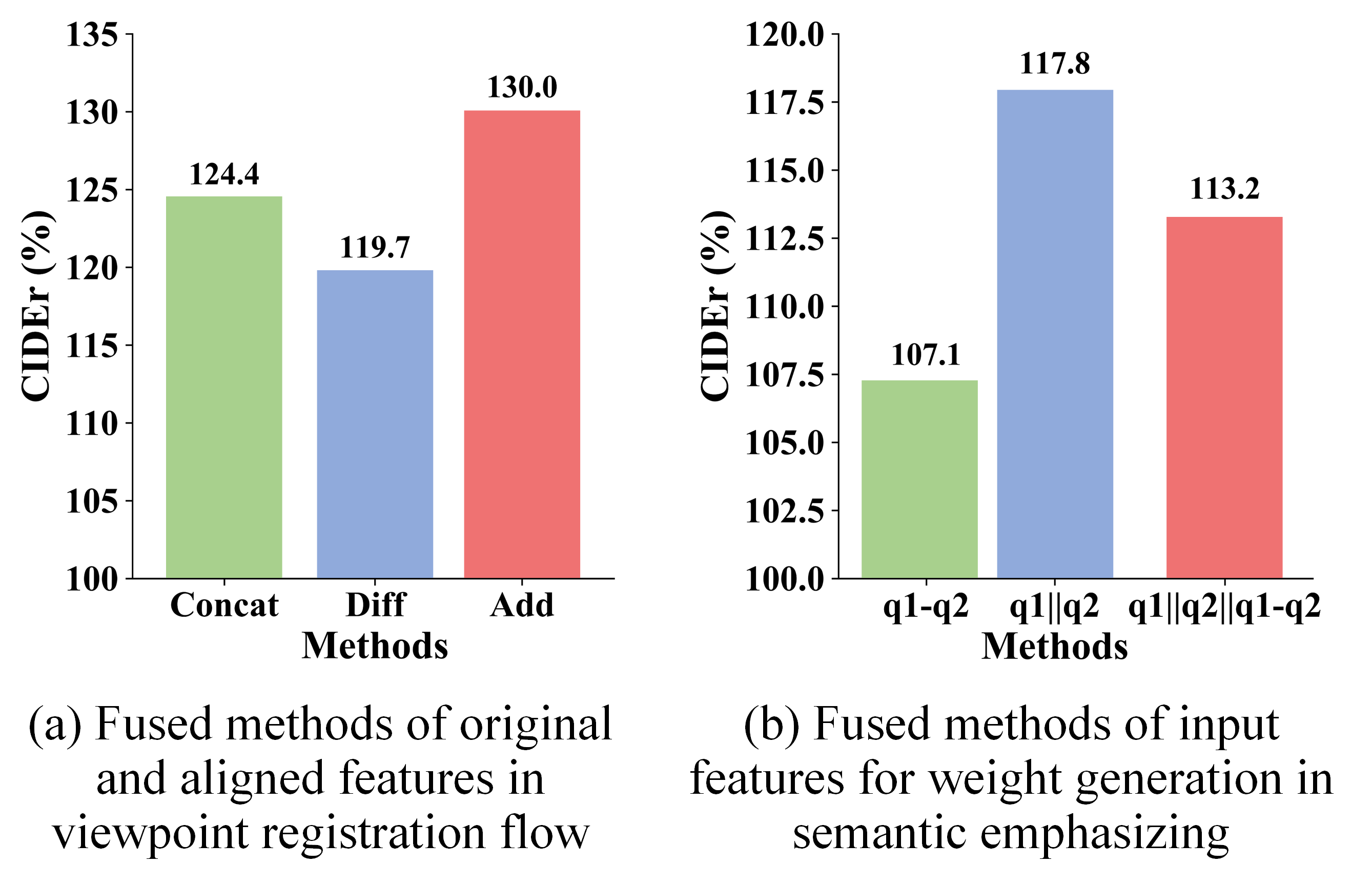} % Reduce the figure size so that it is slightly narrower than the column.
\caption{Ablation study of fused methods of (a) original and aligned features in viewpoint registration flow and (b) features for weight generation in semantic emphasizing.}
\label{fig6}
\end{figure}

\subsubsection{Ablation of Fused Adapter Image Encoder.}

We conduct an ablation on the number of interval transformer blocks for fused adapters, $N$, in the fused adapter image encoder. As shown in Table~\ref{Table 2}, there is a significant improvement in performance when only inserting adapters, indicating the necessity of using adapters to fine-tune the pre-trained encoder. Subsequently, we insert fused adapters and gradually reduce $N$. It is found that $N=5$ results in the best performance. Further reduction in $N$ leads to a slight decrease in performance, indicating the importance of intervals between adapters and fused adapters to separately extract internal image features and integrate inter-image nuance features. Consequently, we set $N=5$ for the remaining experiments.

\subsubsection{Ablation of Viewpoint Registration Flow.}

Figure~\ref{fig5} depicts the predicted viewpoint flow fields under scene changes and distractors. It can be observed that the predicted flow fields can effectively eliminate viewpoint variations between images, and differentiate changed regions by predicting viewpoint flows that differ from the regions with only viewpoint variations. More visualization of the flow fields can be found in the supplementary material.

Besides, directly subtracting viewpoint-aligned features leads to a loss of contextual information regarding the changed objects, making the model only aware of the changed objects without the state before and after the change. We conduct an ablation study on three fused methods including concatenation, subtraction, and addition. The CIDEr results are presented in Figure~\ref{fig6}(a). As expected, using the addition operation results in the best performance.

\subsubsection{Ablation of Semantic Emphasizing.}

To explore the effects of fused methods of input features for weight generation in semantic emphasizing, we select three fused methods, including $q_{1}-q_{2}$, $\left [ q_{1}||q_{2} \right ]$, and $\left [ q_{1}||q_{2}||q_{1}-q_{2} \right ]$. The CIDEr results are shown in Figure~\ref{fig6}(b). $\left [ q_{1}||q_{2} \right ]$ achieves the best performance. For $q_{1}-q_{2}$, directly using them or concatenating them with the original features both result in decreased performance. Thus, $\left [ q_{1}||q_{2} \right ]$ is applied in the semantic emphasizing module.

\begin{table}[t]
	\setlength{\tabcolsep}{5.0pt}
	\renewcommand{\arraystretch}{1.2}
	\begin{center}
		\begin{tabular}{c|ccccc}
			\toprule
			Method & B & C & M & S & R \\ 
			\hline
			Capt-Dual-Att \shortcite{park2019robust} & 43.5 & 108.5 & 32.7 & 23.4 & - \\
                DUDA \shortcite{park2019robust} & 47.3 & 112.3 & 33.9 & 24.5 & - \\
                M-VAM \shortcite{shi2020finding} & 50.3 & 114.9 & 37.0 & 30.5 & 69.7 \\
                M-VAM+ \shortcite{shi2020finding} & 51.3 & 115.8 & 37.8 & 30.7 & 70.4 \\
                IFDC \shortcite{huang2021image} & 49.2 & 118.7 & 32.5 & - & 69.1 \\
                DUDA+Aux \shortcite{hosseinzadeh2021image} & 51.2 & 115.4 & 37.7 & 31.1 & 70.5 \\
                VACC \shortcite{kim2021agnostic} & 52.4 & 114.2 & 37.5 & 31.0 & - \\
                SRDRL \shortcite{tu2021semantic} & 54.9 & 122.2 & 40.2 & 32.9 & 73.3 \\
                $\rm R^3Net$ \shortcite{tu2021r} & 54.7 & 123.0 & 39.8 & 32.6 & 73.1 \\
                BiDiff \shortcite{sun2022bidirectional} & 54.2 & 118.1 & 38.3 & 31.7 & - \\
                IDC-PCL \shortcite{yao2022image} & 51.2 & 128.9 & 36.2 & - & 71.7 \\
                CLIP4IDC \shortcite{guo2022clip4idc} & 56.9 & 150.7 & 38.4 & - & 76.4 \\
                NCT \shortcite{tu2023neighborhood} & 55.1 & 124.1 & 40.2 & 32.9 & 73.8 \\
			\hline
			\textbf{VIR-VLFM (ours)} & \textbf{58.2} & \textbf{153.4} & \textbf{42.6}  & \textbf{34.5} & \textbf{78.9} \\ 
			\bottomrule
		\end{tabular}
		\caption{Change caption performance on CLEVR-Change.}
		\label{Table clevr}
	\end{center}
\end{table}

\begin{table}[!t]
	\setlength{\tabcolsep}{5.0pt}
	\renewcommand{\arraystretch}{1.2}
	\begin{center}
		\begin{tabular}{c|ccccc}
			\toprule
			Method & B & C & M & S & R \\ 
			\hline
			DDLA \shortcite{jhamtani2018learning} & 8.1 & 34.0 & 11.5 & - & 28.3 \\
                DUDA \shortcite{park2019robust} & 8.1 & 32.5 & 11.8 & - & 29.1 \\
                M-VAM \shortcite{shi2020finding} & 10.1 & 38.1 & 12.4 & - & 31.3 \\
                M-VAM+ \shortcite{shi2020finding} & 11.1 & 42.5 & 12.9 & 17.1 & 33.2 \\
                IFDC \shortcite{huang2021image} & 8.7 & 37.0 & 11.7 & - & 30.2 \\
                DUDA+Aux \shortcite{hosseinzadeh2021image} & 8.1 & 34.5 & 12.5 & - & 29.9 \\
                VACC \shortcite{kim2021agnostic} & 9.7 & 41.5 & 12.6 & - & 32.1 \\
                SRDRL \shortcite{tu2021semantic} & - & 35.3 & 13.0 & 18.0 & 31.0 \\
                $\rm R^3Net$ \shortcite{tu2021r} & - & 36.6 & 13.1 & 18.8 & 32.6 \\
                BiDiff \shortcite{sun2022bidirectional} & 6.6 & 42.2 & 10.6 & - & 29.5 \\
                CLIP4IDC \shortcite{guo2022clip4idc} & 11.6 & 47.4 & 14.2 & - & 35.0 \\
			\hline
			\textbf{VIR-VLFM (ours)} & \textbf{12.2} & \textbf{48.9} & \textbf{15.3}  & \textbf{20.1} & \textbf{36.2} \\ 
			\bottomrule
		\end{tabular}
		\caption{Change caption performance on Spot-the-Diff.}
		\label{Table spot}
	\end{center}
\end{table}

\begin{figure}[t]
\centering
\includegraphics[width=\columnwidth]{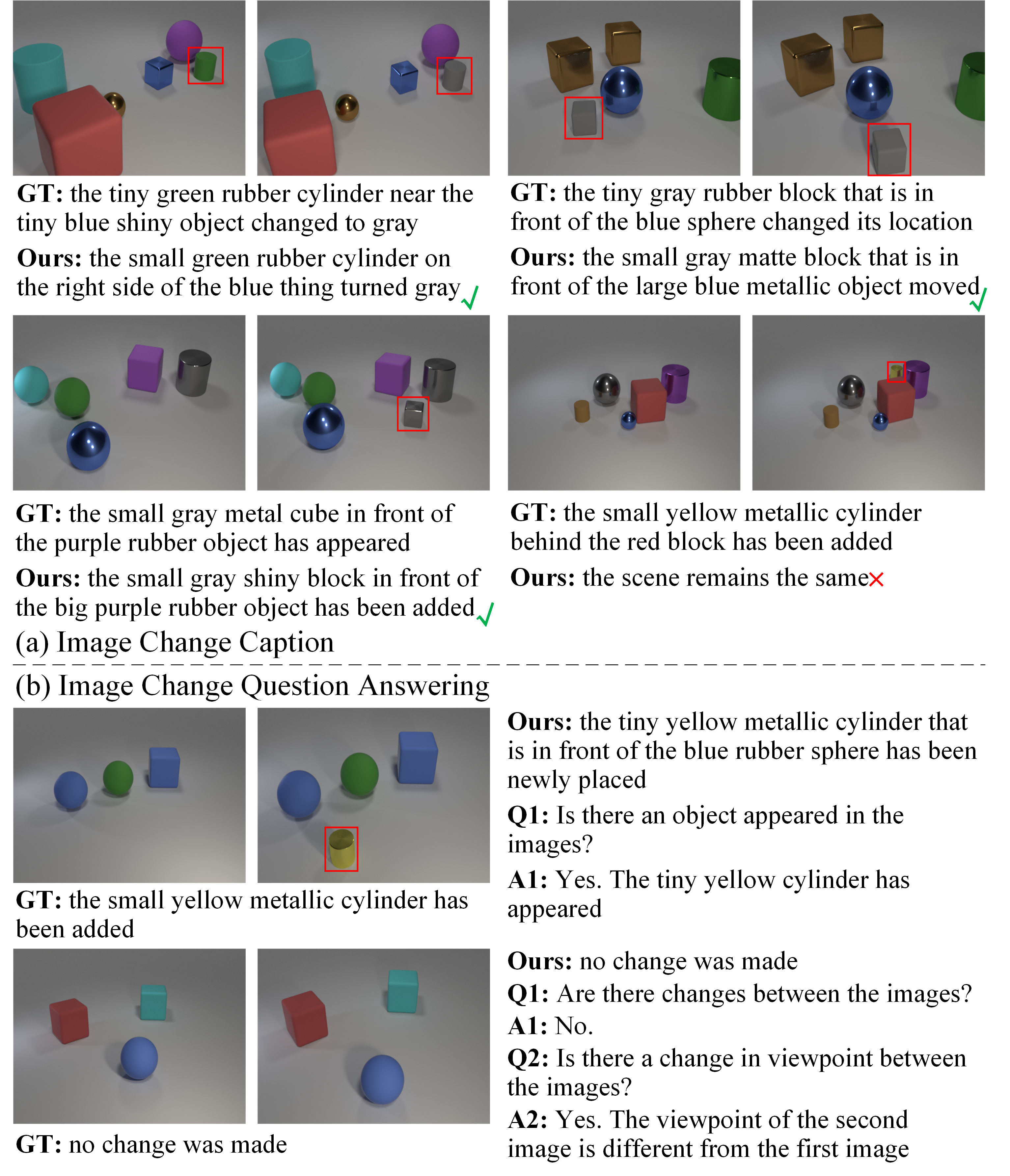} % Reduce the figure size so that it is slightly narrower than the column.
\caption{Qualitative results of our method for (a) image change caption and (b) image change question answering.}
\label{fig7}
\end{figure}

\subsection{Comparisons with State-of-the-art Methods}

We conduct experiments on CLEVR-Change and Spot-the-Diff, and compare our method with state-of-the-art methods.

\subsubsection{CLEVR-Change.}

Table~\ref{Table clevr} presents the experimental results on the CLEVR-Change dataset. Our method outperforms all existing approaches across all metrics, including CLIP4IDC \shortcite{guo2022clip4idc} which also employs a pre-trained foundation model. This indicates that our method effectively enhances VLFMs to comprehend changes between images and mitigate the interference of viewpoint variations.

We present some examples of image change caption in Figure~\ref{fig7}(a). Our model can accurately describe the changes between images in most scenarios, encompassing color changes, position movements, and object appearance or disappearance. While some incorrect descriptions may arise when changes pertain to objects with pronounced occlusion. Furthermore, it is surprising that even without training on question answering data, our model demonstrates competency in question answering tasks in simple scenarios, as Figure~\ref{fig7}(b) shows. We believe that with the incorporation of more data, our model has the potential to acquire a robust multi-image comprehension ability, effectively accomplishing tasks involving change caption and question answering.

\subsubsection{Spot-the-Diff.}

We also evaluate our model on a realistic dataset Spot-the-Diff. As Table~\ref{Table spot} shows, our method still achieves state-of-the-art performance on all metrics. It demonstrates that our VIR-VLFM is generalizable and achieves splendid performance in real-world scenarios.

\section{Conclusion}

In this work, we propose a viewpoint integration and registration method to equip VLFMs with multi-image understanding ability for ICU. A fused adapter image encoder is devised to fine-tune the pre-trained image encoder to capture nuances between images by inserting adapters and fused adapters. Besides, we propose a viewpoint registration flow and a semantic emphasizing module to reduce performance degradation caused by viewpoint variations. Extensive experiments validate the effectiveness of our method which achieves state-of-the-art performance in all metrics. In the future, we intend to further enhance multi-image understanding capability of VLFMs by collecting extensive data and refining model architecture to tackle complex scenarios.

\bibliography{aaai24}

% \newpage
\appendix
\section{Supplementary Material}
In the supplementary material, we provide additional materials to supplement our main submission. Firstly, we present detailed introductions of the experimental setup, including the selected datasets, evaluation metrics, and training details. Additionally, we further showcase some qualitative results to demonstrate the effectiveness of our proposed model, including the predicted viewpoint flow fields as well as the predicted results for image change caption and image change question answering tasks.

\subsection{Experimental Setup}

\subsubsection{Dataset.}

Similar to the majority of previous methods, our method is evaluated on two standard benchmark datasets for image change caption, including CLEVR-Change dataset \cite{park2019robust} and Spot-the-Diff dataset \cite{jhamtani2018learning}.
CLEVR-Change is a synthetic dataset built upon the CLEVR engine \cite{johnson2017clevr}. It portrays scene changes of a set of basic geometry objects with moderate viewpoint change. For each default \textless before\textgreater image, two \textless after\textgreater images are constructed. One depicts changes induced solely by factors like alterations in camera position/ zoom or illumination, serving as a distractor. The other displays actual scene changes of 5 types, consisting of color variation, texture change, new object appearance, object disappearance, and object movement. Each pair of images are annotated with multiple sentences. The dataset comprises 39,803 \textless before\textgreater images and 79,606 \textless after\textgreater images with 493,735 captions. We use the official dataset split with 67,660, 3,976, and 7,970 image pairs for training, validation, and test, respectively.
Spot-the-Diff is a realistic dataset with no distractor. It is constructed based on real image frames extracted from the VIRAT surveillance video dataset with 329 videos across 11 frames of reference \cite{oh2011large}. Each image pair features one or more changes. The annotations are created by crowd-sourcing natural language descriptions of differences between images using Amazon Mechanical Turk. On average, there are 1.86 reported sentences per image pair. The dataset contains 13,192 image pairs. We also use the official split with a ratio of 8:1:1 for training, validation, and test, respectively.

\subsubsection{Evaluation Metrics.}

Following previous works, five standard metrics for image caption task are applied to evaluate the quality of predicted sentences on the test split, including BLEU-4 (B) \cite{papineni2002bleu}, METEOR (M) \cite{banerjee2005meteor}, CIDEr (C) \cite{vedantam2015cider}, SPICE (S) \cite{anderson2016spice}, and ROUGE-L (R) \cite{lin2004rouge}. For inference, we employ a vocabulary ranking method following InstructBLIP \cite{dai2023instructblip}. Specifically, we prompt the model to generate change descriptions within a limited list of candidate words, and then calculate log-likelihood of each candidate word. Finally, the candidate with the highest value is selected as the final prediction. Table~\ref{Table 1-param} presents the detailed hyper-parameters for inference. The results in our experiments are reported based on the Microsoft COCO evaluation server \cite{chen2015microsoft}.

\begin{table}[t]
	\setlength{\tabcolsep}{0.9pt}
	\renewcommand{\arraystretch}{1.2}
	\begin{center}
		\begin{tabular}{c|c|cc}
			\toprule
			Stage & Hyper-parameter & CLEVR-Change & Spot-the-Diff \\ 
			\hline
                & Epoch & 40 & 100 \\
                & Batch size & 16 & 16 \\
                Training & Learning rate & 5e-5 & 5e-5 \\
                & Min lr & 1e-5 & 1e-5 \\
                & Weight decay & 0.05 & 0.05 \\
                & Warmup steps & 4500 & 650 \\
                \hline
		       & Beams number & 5 & 5 \\
                 & Temperature & 1 & 1 \\
                 Inference & Repetition penalty & 1.5 & 1.5 \\
                 & Word list number & 72 & 2296 \\
                 & Max length & 50 & 100 \\
			\bottomrule
		\end{tabular}
		\caption{Hyper-parameters used for training and inference.}
		\label{Table 1-param}
	\end{center}
\end{table}
\vspace{-1em}

\begin{figure*}[t]
\centering
\includegraphics[width=\linewidth]{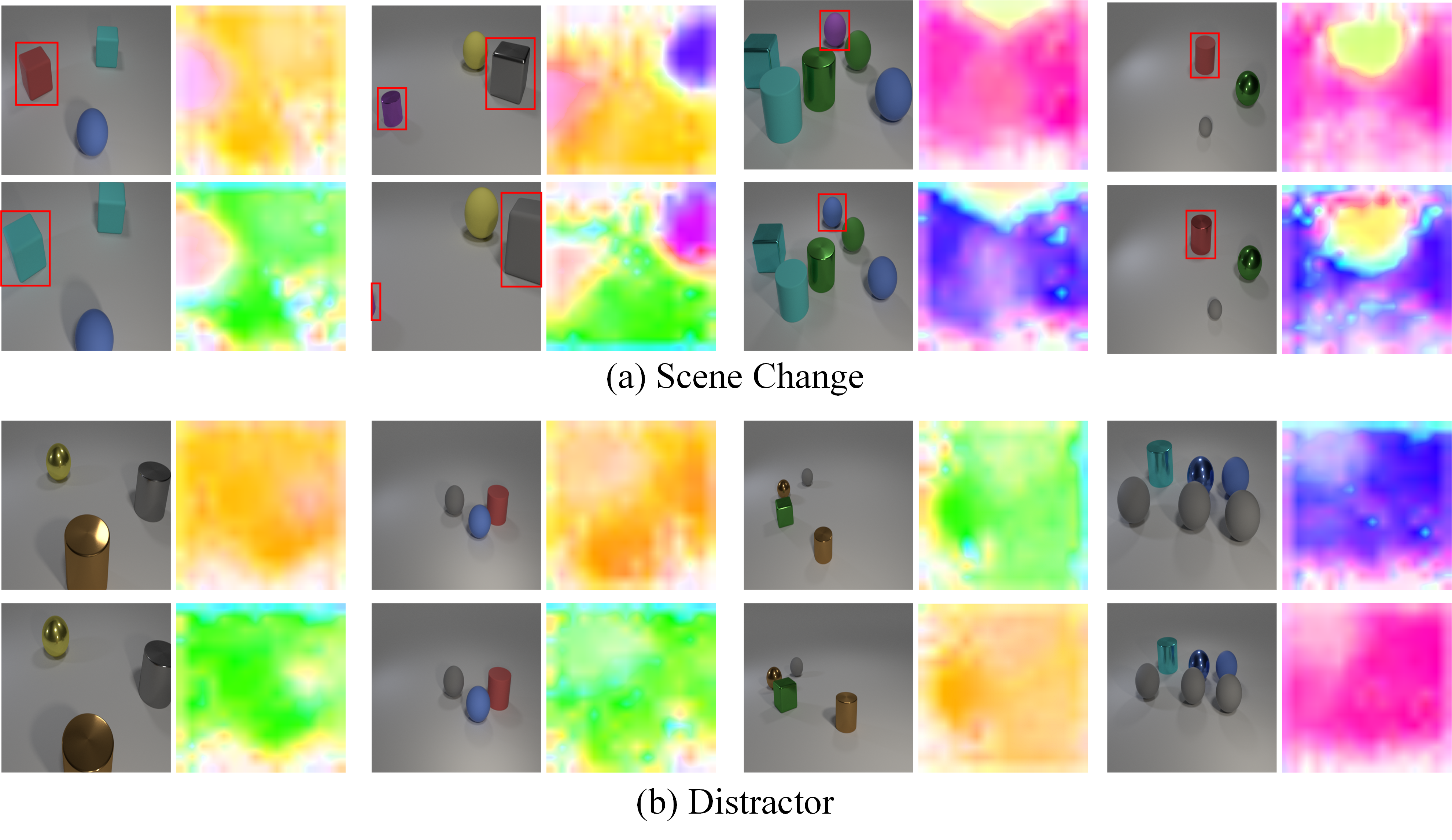} % Reduce the figure size so that it is slightly narrower than the column.
\caption{Visualization of viewpoint registration flow fields.}
\label{sup-fig1}
\end{figure*}
\vspace{-1em}

\subsubsection{Implementation Details.}

The architecture of our model is roughly consistent with BLIP-2 \cite{li2023blip} and InstructBLIP \cite{dai2023instructblip}. We apply EVA-ViT-g/14 \cite{fang2023eva} and Vicuna-7B \cite{chiang2023vicuna} as the image encoder and the large language model, respectively. The two modules are augmented with a Q-Former and a fully connected layer pre-trained by BLIP-2. The network is optimized by using the AdamW optimizer \cite{loshchilov2017decoupled} with $\beta =\left ( 0.9, 0.99 \right ) $ and the weight decay of 0.05. We set batch size and learning rate as 16 and 5e-5, respectively. Consistent with DUDA \cite{park2019robust}, we train our model for 40 and 100 epochs on the CLEVR-Change and Spot-the-Diff dataset, respectively. The training details of the two datasets are presented in Table~\ref{Table 1-param}. We implement our experiments in PyTorch \cite{paszke2019pytorch} framework on 8 V100 GPUs.

\subsection{Qualitative Analysis}
In this section, we present qualitative analysis of our proposed model on the CLEVR-Change dataset, including the predicted flow fields in the devised viewpoint registration flow and some examples of our model in image change caption and question answering tasks.

\subsubsection{Visualization of Viewpoint Registration Flow Fields.}

Figure~\ref{sup-fig1} presents additional visualizations of the predicted viewpoint flow fields in the viewpoint registration flow under both scene changes and distractors. The results further illustrates that our designed viewpoint registration flow can effectively predict the viewpoint differences between two images and achieve viewpoint alignment by the warp operation, aiding the subsequent modules in identifying the regions of genuine changes and generating descriptions.

\subsubsection{Qualitative Results of Image Change Caption.}

\begin{figure*}[t]
\centering
\includegraphics[width=\linewidth]{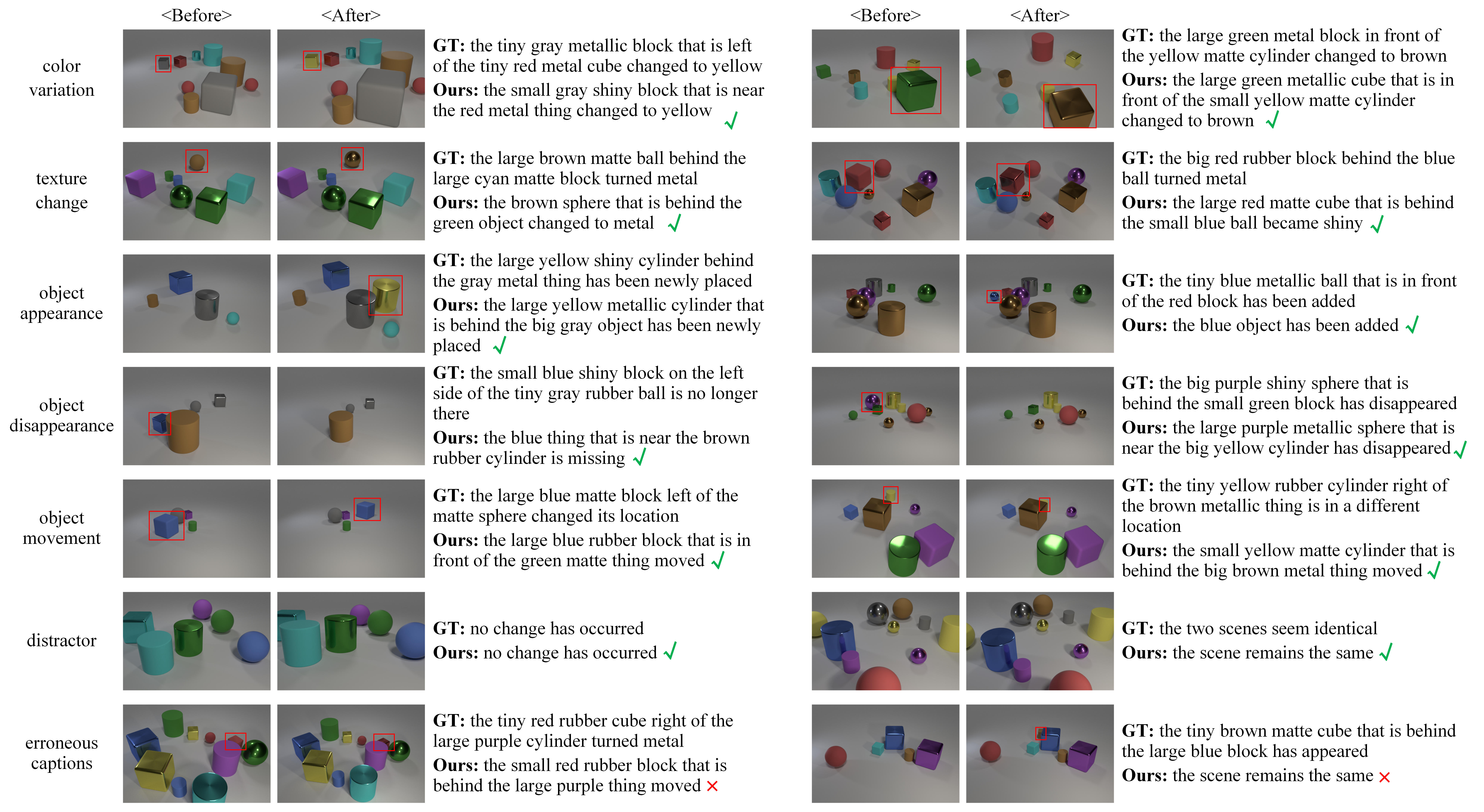} % Reduce the figure size so that it is slightly narrower than the column.
\caption{Qualitative results of our method for image change caption.}
\label{sup-fig2}
\end{figure*}

\begin{figure*}[tb]
\centering
\includegraphics[width=0.8\linewidth]{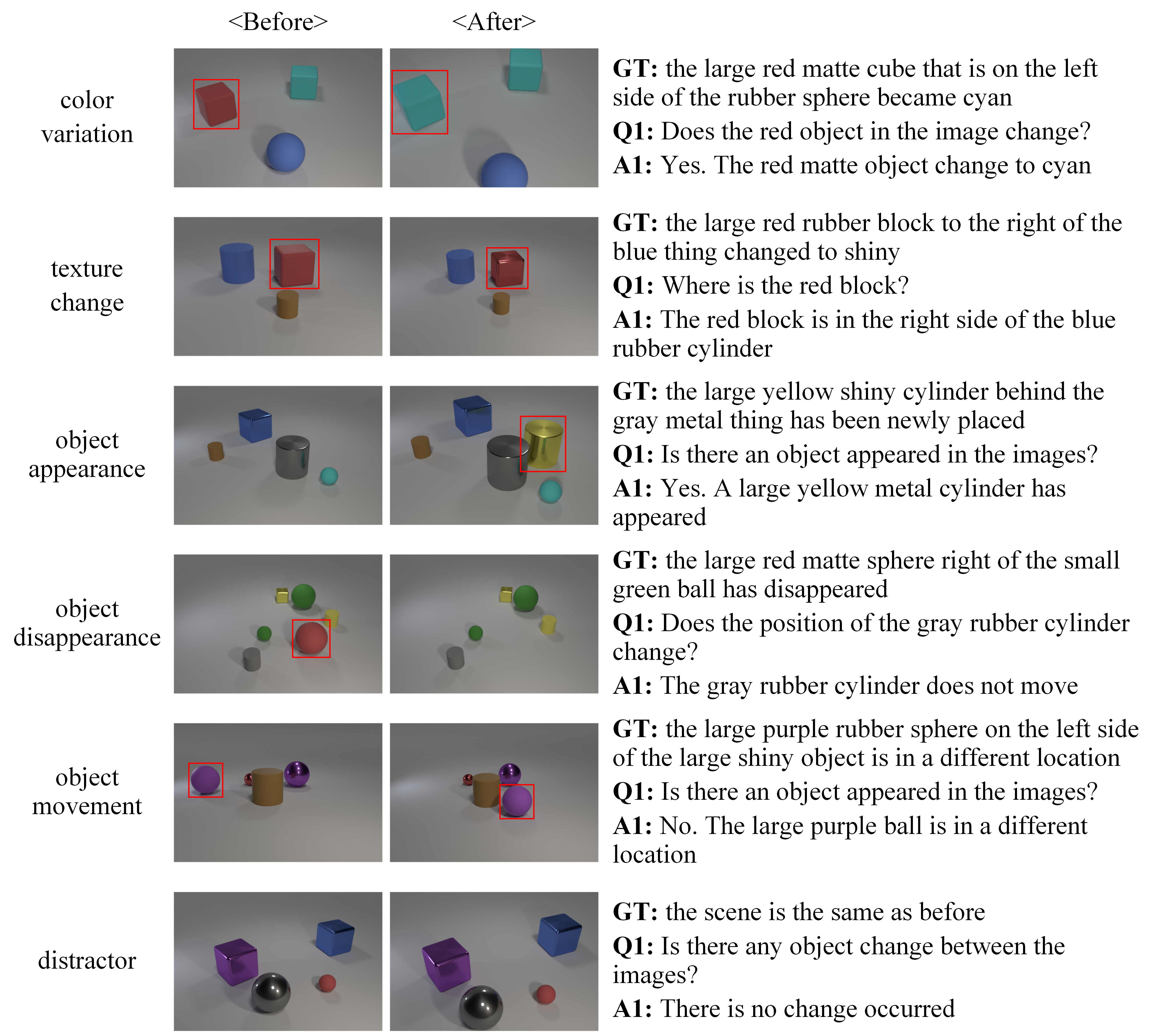} % Reduce the figure size so that it is slightly narrower than the column.
\caption{Qualitative results of our method for image change question answering.}
\label{sup-fig3}
\end{figure*}

In Figure~\ref{sup-fig2}, we present some results of our model in image change caption task. It shows that our approach has the ability to comprehend the correlations between two images and accurately describe the actual changes in most scenarios. But in some situations where changed objects are significantly occluded, there might be instances of erroneous descriptions.

\subsubsection{Qualitative Results of Image Change Question Answering.}

Surprisingly, even without utilizing question answering data during training, our model exhibits preliminary capabilities in image change question answering tasks. As depicted in Figure~\ref{sup-fig3}, in some straightforward scenarios, our model, relying on the generalized knowledge of pre-trained vision language foundation models, can accurately answer some simple questions related to changes. We anticipate that with the integration of more training data, our model will acquire robust
multi-image comprehension ability. In future work, we plan to further explore the robust multi-image understanding ability of vision language foundation models.

\end{document}